\documentclass[letterpaper]{article} 
\usepackage{aaai2027}
\usepackage[hyphens]{url}  
\usepackage{graphicx} 
\usepackage{natbib}  
\usepackage{caption} 
\usepackage{algorithm}
\usepackage{algorithmic}

\usepackage{newfloat}
\usepackage{listings}
\DeclareCaptionStyle{ruled}{labelfont=normalfont,labelsep=colon,strut=off} 
\floatstyle{ruled}
\newfloat{listing}{tb}{lst}{}
\floatname{listing}{Listing}

\usepackage{booktabs}

\usepackage{amsmath}
\usepackage{amssymb}
\usepackage{svg}
\usepackage{subcaption}
\usepackage{graphicx}   
\usepackage{booktabs}   
\usepackage{multirow}   
\usepackage[table]{xcolor} 

\title{EviAnchor: Mitigating Hallucinations in Large Vision-Language Models via Regional Visual Evidence Compensation}

\author{
    Sihang Jia\textsuperscript{\rm 1},
    Shuliang Liu\textsuperscript{\rm 1},
    Songbo Yang\textsuperscript{\rm 1},
    Xuming Hu\textsuperscript{\rm 1}\corresponding
}
\affiliations{
    \textsuperscript{\rm 1}The Hong Kong University of Science and Technology (Guangzhou)\\
    sjia188@connect.hkust-gz.edu.cn
}

\begin{document}

\maketitle

\begin{abstract}
Large vision-language models (LVLMs) frequently generate content unsupported by visual inputs. Preliminary experiments show that visual evidence is primarily incorporated into answer-side representations in early-to-middle decoder layers, while its direct influence progressively weakens in later layers. This attenuation suggests that visual evidence acquired earlier may be insufficiently utilized during subsequent generation. Based on this observation, we propose \textbf{EviAnchor}, a training-free and single-branch inference framework that preserves and reactivates visual evidence throughout generation. EviAnchor introduces Regional Evidence Anchor (REA) slots to progressively aggregate dense visual tokens into spatially structured representations. It then strengthens the current decision state’s access to these visual anchors through decision-conditioned evidence routing, mitigating excessive dependence on textual context. Finally, the model resumes its native Transformer computation to integrate the retrieved visual evidence with question semantics and generation history. Experiments across POPE, CHAIR, and MMHal-Bench demonstrate consistent improvements in visual grounding.
\end{abstract}


\section{Introduction}
\label{sec:introduction}

In recent years, large vision-language models (LVLMs) have achieved remarkable progress by integrating visual encoders with large language models. Nevertheless, they still frequently generate objects, attributes, or relationships unsupported by the input image, known as \textbf{multimodal hallucination}~\cite{li2023evaluating,liu2024survey,bai2024hallucination,guan2024hallusionbench}, which limits their reliability in real-world applications.

Due to the considerable computational cost of training-based approaches~\cite{sun2024aligning,liu2024mitigating,hu2025prescribing,lu2025mitigating}, training-free mitigation has received increasing attention. Existing methods mainly intervene at two levels.
\emph{Output-space methods} revise token selection through contrastive distributions or candidate-level penalties~\cite{leng2024mitigating,huang2024opera,zhao2025crossimagecontrastivedecodingprecise,zhu2025ibd,chen2024halc,an2025mitigating}.
Although they directly control token selection without modifying the backbone parameters, they primarily intervene after cross-modal representations have already been formed and therefore make it difficult to explicitly exploit how visual evidence propagates. Some contrastive variants additionally require multiple decoding branches~\cite{leng2024mitigating,zhao2025crossimagecontrastivedecodingprecise}. 
\emph{Representation-space methods} instead intervene in hidden states,
attention patterns, parameter subspaces, or internal memory~\cite{liu2025reducing,yin2025clearsight,yang2025nullu,zou2024look,tang2025seeing,su2025activation,wang2025shift,chen2025ict}. While providing more direct access to internal computation, these methods generally lack an explicit mechanism for preserving spatially structured visual evidence across layers and retrieving region-specific evidence conditioned on the current generation state.

In this work, we revisit multimodal hallucination from the perspective of the hierarchical computation process inside LVLMs. A LVLM containing $L$ Transformer decoder layers does not process visual and textual information uniformly
across depth~\cite{chuang2024dola,jiang2025devils,song2026does,kaduri2025s}.

\begin{figure*}[t]
  \centering

  \begin{subfigure}[t]{0.45\linewidth}
      \centering
      \includegraphics[width=\linewidth]{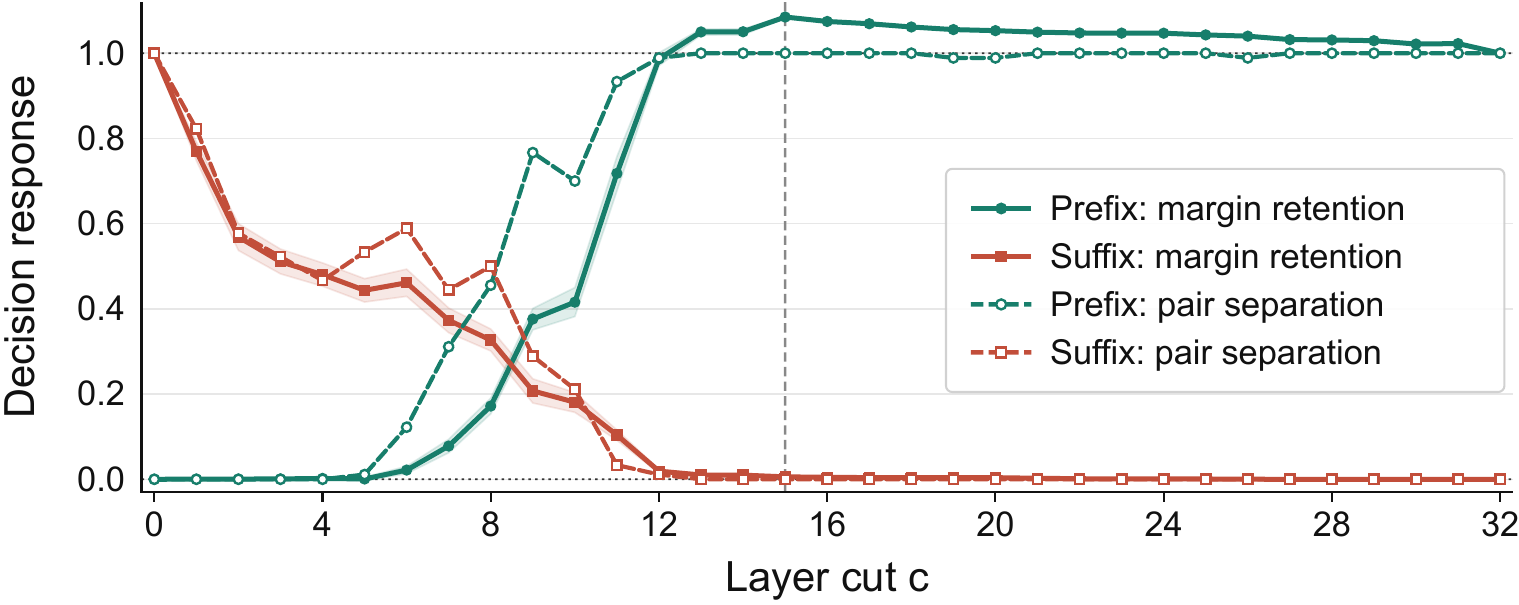}
      \caption{Causal localization of visual-to-text information transfer.}
      \label{fig:path_scan}
  \end{subfigure}
  \begin{subfigure}[t]{0.45\linewidth}
      \centering
      \includegraphics[width=\linewidth]{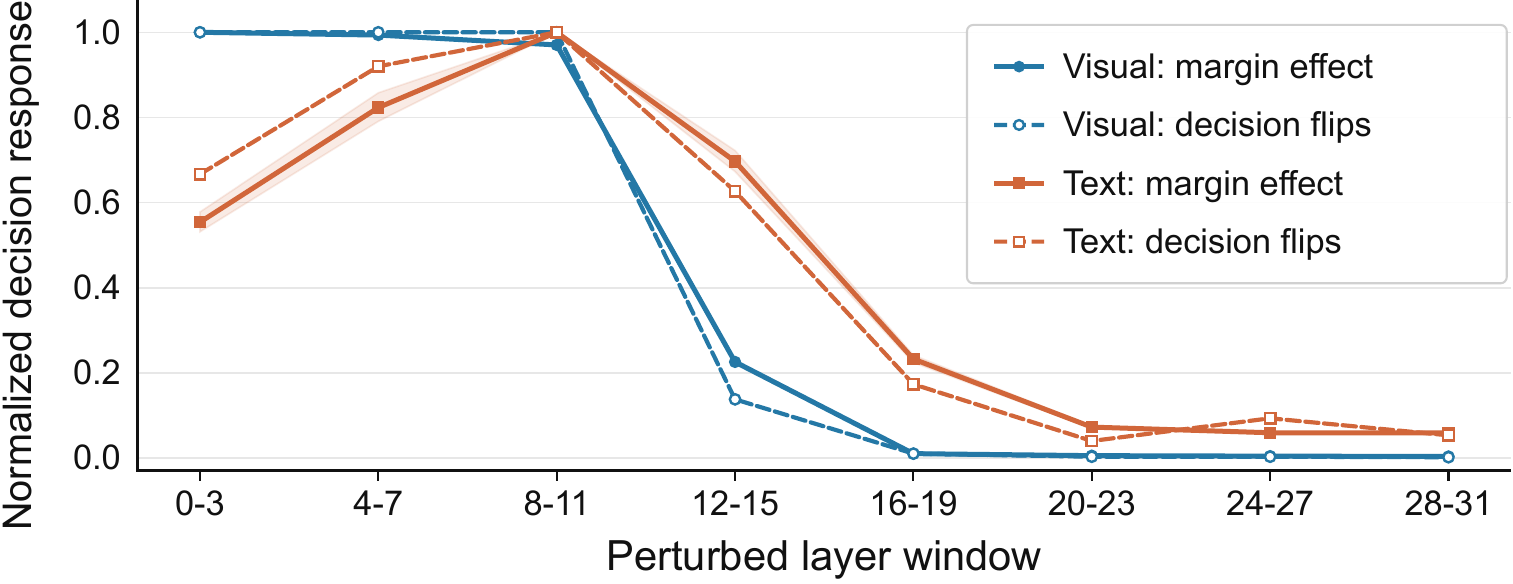}
      \caption{Layer-wise counterfactual sensitivity.}
      \label{fig:vt_counterfactual}
  \end{subfigure}
  \caption{
    Layer-wise causal analysis of visual information flow.
    \textbf{(a)} We pair images containing opposite visual facts and selectively retain text-to-visual attention either before or after each layer boundary. The resulting image-conditioned decision gap measures where visual information is transferred into answer-side representations.
    \textbf{(b)} We replace visual-token states within successive layer windows with those from a blank-image branch, while perturbing question-token states as a textual reference, and measure the resulting decision changes.
    Additional implementation details are provided in the supplementary material.}

  \label{fig:layerwise_visual_analysis}
\end{figure*}

We conduct attention-path restriction and layer-wise counterfactual analyses to localize visual influence across the decoder. As shown in Figures~\ref{fig:path_scan} and~\ref{fig:vt_counterfactual}, visual evidence has its strongest effect in the early-to-middle layers, while its direct influence rapidly weakens in later layers. These results indicate that visual information is primarily written into answer-side representations in the first half of the model.

\begin{figure}
    \centering
    \includegraphics[width=1\linewidth]{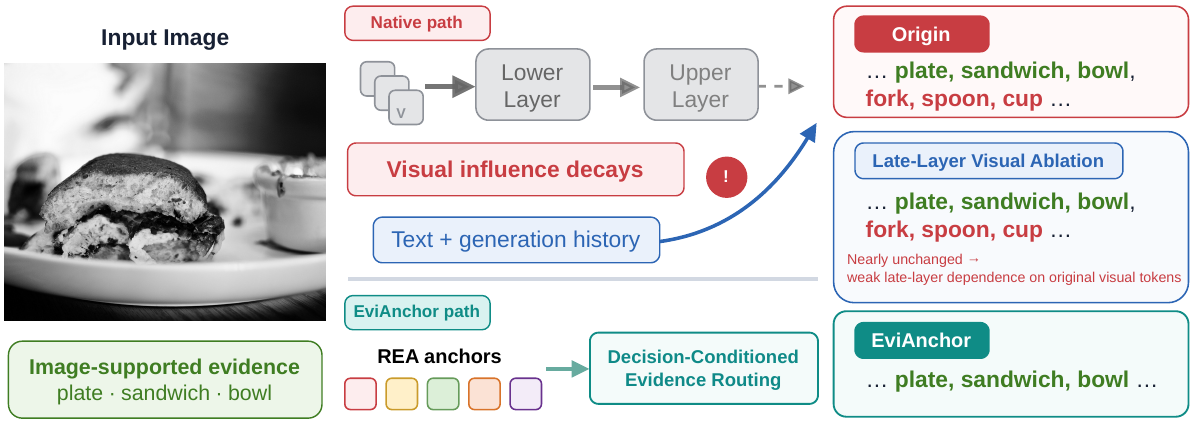}
    \caption{Qualitative illustration of insufficient cross-layer utilization of visual evidence.
    The native model generates unsupported objects highlighted in red.
    Ablating the original visual-token states in later layers leaves these hallucinated mentions nearly unchanged, indicating that the prediction has limited dependence on late-layer visual representations.
    } 
    \label{fig:failure_case}
\end{figure}

These observations reveal an overlooked discrepancy: visual evidence being read in the lower layers does not imply that it will be continuously and reliably used by higher-layer generation states. As the network becomes deeper, the direct influence of the original visual tokens progressively weakens, making the previously acquired evidence more likely to be underutilized during generation. From this perspective, multimodal hallucination can be understood as insufficient cross-layer propagation of visual evidence.
A representative example of this failure mode is provided in Figure~\ref{fig:failure_case}.

Based on this insight, we propose \textbf{EviAnchor}, an inference-time hallucination mitigation method for LVLMs. Unlike contrastive decoding methods, EviAnchor maintains only a single modified multimodal forward path and consists of three functional modules. The Regional Evidence Anchor preserves visual evidence acquired in earlier layers by introducing \emph{regional evidence anchor slots} (REA slots), which provide spatially structured visual representations for subsequent generation states. The Decision-Conditioned Evidence Routing strengthens the answer-side state's access to relevant visual evidence by loosening the attention attraction of the target text and redirects its attention mass to the fixed REA slots. The Native Semantic Recomposition restores the original Transformer computation, allowing the model to integrate visual evidence with question semantics and generation history while maintaining linguistic coherence.

The main contributions of this work are summarized as follows:

\begin{enumerate}
    \item We introduce a regional visual evidence anchor mechanism based on dynamic regional pooling. Without updating model parameters, REA slots continuously aggregate local visual information in the lower layers, producing a compact collection of visual anchors.

    \item We propose a decision-conditioned visual evidence retrieval mechanism that reconnects the generation state with relevant visual evidence. By reducing excessive reliance on textual representations, the mechanism dynamically transfers attention probability mass from target-text tokens to REA slots.

    \item We present EviAnchor, a single-branch and training-free framework requiring no external tools or multi-branch logit subtraction. We evaluate its effectiveness on both discriminative and generative benchmarks.
\end{enumerate}

\begin{figure*}[t]
    \centering
    \includegraphics[width=\textwidth]{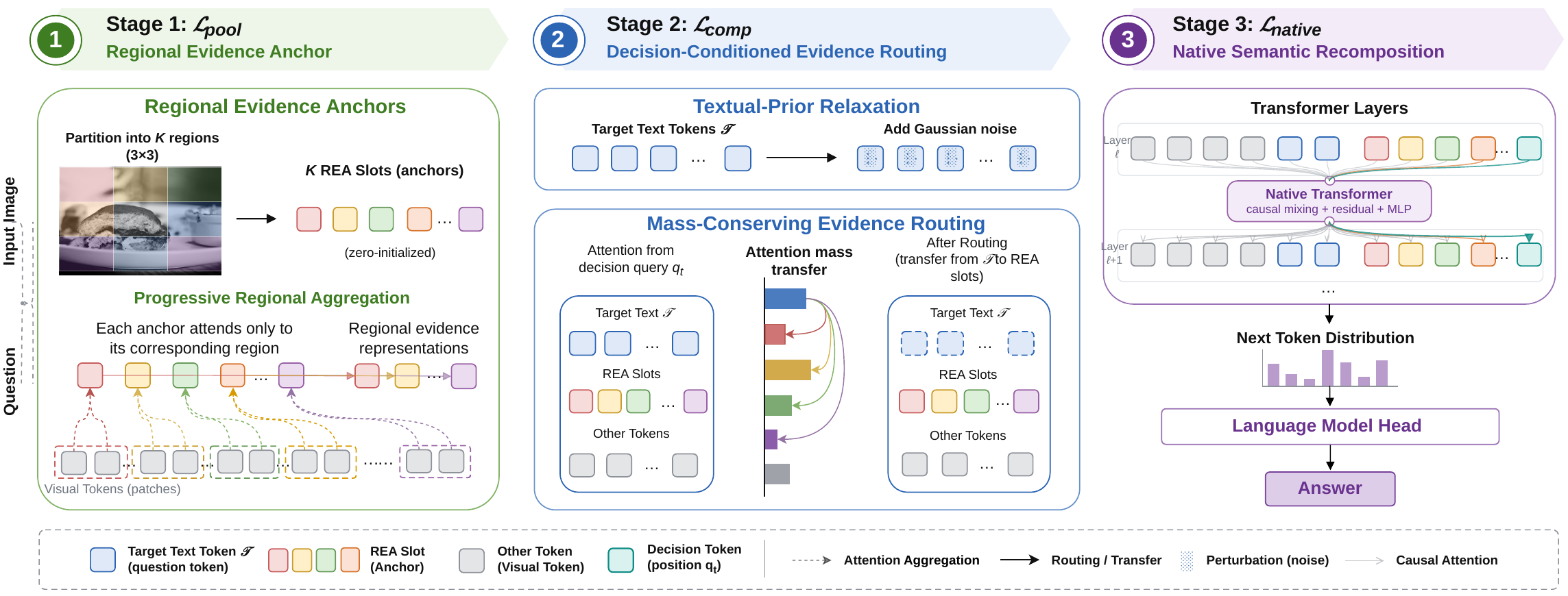}
    \caption{
    Overview of EviAnchor. The Regional Evidence Anchor aggregates visual patches into spatial REA slots, the Decision-Conditioned Evidence Routing reconnects the current generation state with these visual anchors, and the Native Semantic Recomposition restores the original Transformer computation for coherent response generation.
    }
    \label{fig:method_overview}
\end{figure*}

\section{Related Work}
\label{sec:related_work}

\subsection{Multimodal Hallucination Mitigation}

Large vision-language models (LVLMs) frequently generate objects, attributes, or relationships that are unsupported by visual inputs, which is commonly referred to as multimodal hallucination~\cite{liu2024survey,fu2024mme}. Existing analyses attribute hallucination to insufficient visual grounding, over-reliance on language priors, and ineffective utilization of visual representations~\cite{li2023evaluating,gunjal2024detecting,bai2024hallucination,guan2024hallusionbench}. Recent mechanistic studies further reveal that multimodal information is processed unevenly across Transformer layers: early layers mainly incorporate visual signals, while deeper layers gradually shift toward language-dominated semantic processing~\cite{jiang2025devils,song2026does,kaduri2025s}. 

Existing mitigation approaches can be broadly divided into training-based~\cite{sun2024aligning,liu2024mitigating,hu2025prescribing,lu2025mitigating} and inference-time methods. However, training-based approaches require additional data and optimization, limiting their applicability to existing LVLMs. Therefore, recent studies increasingly focus on inference-time mitigation.

\subsection{Output-Space Intervention}

Output-space methods~\cite{leng2024mitigating,wang2024mitigating,huang2024opera,jia2026decodingperturbationmitigatingmllm,liu2026visionlanguageintrospectionmitigatingoverconfident} mitigate hallucination by directly modifying token probabilities during decoding. VCD~\cite{leng2024mitigating} introduces a distorted visual branch and contrasts output distributions to reduce the influence of language priors. Following this direction, ICD~\cite{wang2024mitigating} incorporates instruction-aware contrastive signals, while OPERA~\cite{huang2024opera} suppresses hallucination through attention-based decoding penalties. CICD~\cite{zhao2025crossimagecontrastivedecodingprecise} further improves contrastive decoding by exploiting alternative visual conditions.
Although these methods are effective and easy to deploy, they intervene after multimodal representations have already been formed. Consequently, they mainly adjust final predictions rather than explicitly controlling how visual evidence is acquired inside the model~\cite{liu2024survey}. 

\subsection{Representation-Space Intervention}

Representation-space methods~\cite{liu2025reducing,yin2025clearsight,yang2025nullu,zou2024look} directly modify hidden states or internal attention computation to influence the generation trajectory. VTI~\cite{liu2025reducing} adjusts intermediate representations according to visual token importance to enhance visual reliance. ClearSight~\cite{yin2025clearsight} calibrates internal attention patterns to improve visual grounding, while Nullu~\cite{yang2025nullu} explores null-space representation editing to suppress hallucination-related components. More recently, MemVR~\cite{zou2024look} introduces a visual memory reconstruction mechanism by reinjecting visual prompts into FFN key-value memory when visual information becomes insufficient. Different from these approaches, EviAnchor explicitly preserves spatially structured visual evidence through regional anchors during early-layer processing and performs decision-conditioned retrieval of relevant evidence during later generation.
Compared with output-space methods, these approaches provide more direct access to the model's internal computation. However, existing methods mainly focus on modifying hidden representations, calibrating attention, or recovering degraded visual memory. They do not explicitly maintain a layer-aware, spatially organized evidence representation that can be selectively accessed by the current generation state.

\section{Methodology}
\label{sec:methodology}

We propose \textbf{EviAnchor}, a training-free framework for regional visual evidence compensation that operates exclusively at inference time. EviAnchor targets hallucinations associated with the insufficient influence of visual evidence acquired in earlier layers on subsequent token predictions. 

Our layer-wise analysis indicates that visual information acquisition and semantic convergence occur at different stages of generation. Accordingly, we divide the inference process into three consecutive layer intervals:
\begin{itemize}
    \item $\mathcal{L}_{\mathrm{pool}}$ corresponds to the stage in which visual responses are strong and continuously enter the model states;
    \item $\mathcal{L}_{\mathrm{comp}}$ corresponds to the transition stage in which the direct influence of visual tokens begins to weaken and the current decision becomes increasingly mediated by textual hidden states;
    \item $\mathcal{L}_{\mathrm{native}}$ corresponds to the stage in which the model organizes the available information into lexical predictions.
\end{itemize}

An overview of the proposed framework is shown in Figure~\ref{fig:method_overview}. EviAnchor consists of three functional modules operating over different decoder stages. 

\subsection{Regional Evidence Anchor}
\label{sec:regional_anchor}

This module provides a persistent interface between the dense visual-token sequence and answer-side representations. Directly enhancing all visual tokens cannot distinguish task-relevant regions from irrelevant backgrounds. Conversely, compressing the entire image into a single global vector may discard fine-grained visual details. Therefore, EviAnchor adopts a regional representation that lies between dense patch-level features and a global image summary.

Before the answer position, we insert $K$ zero-initialized regional evidence anchor slots:
\begin{equation}
E^{(0)}
=
\left\{
e_1^{(0)},
\ldots,
e_K^{(0)}
\right\},
\qquad
e_k^{(0)}
=
\mathbf{0},
\label{eq:rea_initialization}
\end{equation}
and construct the initial multimodal sequence as
\begin{equation}
H^{(0)}
=
\left[
V;
X;
E^{(0)};
p_0
\right],
\label{eq:initial_sequence}
\end{equation}
where $V$ and $X$ denote the visual and textual token sequences, respectively, and $p_0$ denotes the answer-start prediction position.

REA slots are not additionally trained model parameters. Instead, they are temporary hidden positions introduced for the current input. Meanwhile, we partition the spatial grid of visual tokens into $K$ mutually disjoint regions:
\begin{equation}
\mathcal{V}
=
\bigcup_{k=1}^{K}
\mathcal{R}_k,
\qquad
\mathcal{R}_i
\cap
\mathcal{R}_j
=
\varnothing,
\quad
i\neq j.
\label{eq:region_partition}
\end{equation}
The $k$-th REA slot $e_k$ is associated with the corresponding region $\mathcal{R}_k$.

Each REA slot is allowed to attend only to itself and the visual tokens within its corresponding region. Let $M^{(\ell)}$ denote the attention mask at layer $\ell$. The mask associated with the $k$-th REA slot is defined as
\begin{equation}
M_{e_k,j}^{(\ell)}
=
\begin{cases}
0,
&
j
\in
\mathcal{R}_k
\cup
\{e_k\},
\\
-\infty,
&
\text{otherwise},
\end{cases}
\qquad
\ell
\in
\mathcal{L}_{\mathrm{pool}}.
\label{eq:rea_pooling_mask}
\end{equation}

At layer $\ell$ and attention head $h$, the aggregation weight assigned by the $k$-th REA slot to a visual token within its corresponding region is
\begin{equation}
\alpha_{k,j}^{(\ell,h)}
=
\frac{
\exp\left(
Q_{e_k}^{(\ell,h)}
K_j^{(\ell,h)\top}
/\sqrt{d}
\right)
}{
\displaystyle
\sum_{r\in\mathcal{R}_k\cup\{e_k\}}
\exp\left(
Q_{e_k}^{(\ell,h)}
K_r^{(\ell,h)\top}
/\sqrt{d}
\right)
},
\label{eq:regional_attention_weight}
\end{equation}
and the corresponding attention output is
\begin{equation}
o_{e_k}^{(\ell,h)}
=
\sum_{j\in\mathcal{R}_k\cup\{e_k\}}
\alpha_{k,j}^{(\ell,h)}
V_j^{(\ell,h)}.
\label{eq:regional_attention_output}
\end{equation}




Since REA slots are initialized as zero vectors, their early updates provide coarse regional aggregation, while subsequent layers dynamically refine the aggregated evidence through updated queries. To prevent premature interference with generation, this module restricts answer-side positions from accessing REA slots until stable regional representations have been formed.. After this stage, dense visual tokens are transformed into $K$ spatially indexed regional evidence representations.

\subsection{Decision-Conditioned Evidence Routing}
\label{sec:visual_compensation}

The existence of REA slots does not necessarily imply that the current generation state will use them. During later computation, the decision position $q_t$ may allocate most of its attention to the linguistic context, leaving the REA slots available but unread.

To address this issue, the Decision-Conditioned Evidence Router reduces the direct dependency of the current decision on the target text and then reallocates the released attention probability mass to REA slots selected according to the decision state's own regional preference.

\paragraph{Textual-prior relaxation.}

If the target instruction remains a highly stable attention source, the decision position may continue to rely on question templates and linguistic co-occurrence patterns while ignoring the regional visual evidence formed in earlier layers.

For the target text-token set $\mathcal{T}$, this module applies scale-controlled Gaussian perturbations to its hidden states within the compensation window:
\begin{equation}
\widehat{h}_j^{(\ell)}
=
h_j^{(\ell)}
+
\lambda
s_\ell
\epsilon_j^{(\ell)},
\qquad
\epsilon_j^{(\ell)}
\sim
\mathcal{N}(0,I),
\label{eq:text_noise}
\end{equation}
where
\begin{equation}
j
\in
\mathcal{T},
\qquad
\ell
\in
\mathcal{L}_{\mathrm{comp}}.
\end{equation}
Here, $s_\ell$ denotes the standard deviation of the target-text hidden states at layer $\ell$, and $\lambda$ controls the perturbation strength.

This operation does not directly enhance visual information. Instead, it weakens the target text as a direct high-level information source, creating an opportunity for the answer state to retrieve evidence from REA slots.

\paragraph{Mass-conserving evidence routing.}

EviAnchor directly modifies the attention distribution of the current decision position $q_t$. Specifically, it transfers probability mass from the target text to the REA slots, thereby injecting regional visual evidence into the current next-token decision.

Let the attention distribution at layer $\ell$ and attention head $h$, after text perturbation but before probability transfer, be
\begin{equation}
A_{t,j}^{(\ell,h)}
=
\operatorname{softmax}_{j}
\left(
\frac{
Q_{q_t}^{(\ell,h)}
K_j^{(\ell,h)\top}
}{
\sqrt{d}
}
+
M_{q_t,j}^{(\ell)}
\right).
\label{eq:original_attention}
\end{equation}

We compute the total attention mass assigned by the current decision state to the target text-token set $\mathcal{T}$:
\begin{equation}
m_t^{(\ell,h)}
=
\sum_{j\in\mathcal{T}}
A_{t,j}^{(\ell,h)}.
\label{eq:text_attention_mass}
\end{equation}

At the same time, the original attention distribution provides the relative preference of $q_t$ for different REA slots. We normalize this preference as
\begin{equation}
\pi_{t,k}^{(\ell,h)}
=
\frac{
\max\left(
A_{t,e_k}^{(\ell,h)},
\varepsilon
\right)
}{
\displaystyle
\sum_{r=1}^{K}
\max\left(
A_{t,e_r}^{(\ell,h)},
\varepsilon
\right)
},
\qquad
\sum_{k=1}^{K}
\pi_{t,k}^{(\ell,h)}
=
1,
\label{eq:regional_preference}
\end{equation}
where $\varepsilon>0$ is a small constant that prevents numerical degeneracy.

EviAnchor reclaims the attention budget assigned to the target text and redistributes it to the REA slots according to the regional preference:
\begin{equation}
\widetilde{A}_{t,j}^{(\ell,h)}
=
\begin{cases}
0,
&
j\in\mathcal{T},
\\[3pt]
A_{t,e_k}^{(\ell,h)}
+
m_t^{(\ell,h)}
\pi_{t,k}^{(\ell,h)},
&
j=e_k,
\\[3pt]
A_{t,j}^{(\ell,h)},
&
\text{otherwise}.
\end{cases}
\label{eq:attention_transfer}
\end{equation}

Because the amount of probability mass removed from the target-text positions is equal to the amount added to the REA slots, the modified attention distribution remains normalized.



The routed attention representation is then propagated through the remaining Transformer computation, allowing the retrieved regional evidence to influence the current decision trajectory. To maintain consistent visual references during routing, the REA slots used as attention sources are kept fixed throughout this module.

\subsection{Native Semantic Recomposition}
\label{sec:native_decoding}

The Decision-Conditioned Evidence Routing has already introduced the retrieved regional evidence into the hidden-state trajectory of the current decision position. However, continuously enforcing textual relaxation and attention redistribution in the upper layers may over-constrain the generation process and impair response coherence. Therefore, Native Semantic Recomposition terminates all explicit routing interventions and delegates the remaining computation to the pretrained model.

The REA slots remain available as regular context positions, allowing the upper-layer decision state to jointly attend to the original visual tokens, the preserved regional evidence, the question, and the generation history. The decision state integrates them through native Transformer computation, and the next-token distribution is produced by the original language-model head:
\begin{equation}
\begin{gathered}
h_{q_t}^{(L)}
=
\operatorname{Transformer}_{\mathcal{L}_{\mathrm{native}}}
\left(
q_t;
\left[
V;
X;
E;
y_{<t}
\right]
\right),
\\[4pt]
p\left(
y_t
\mid
I,X,y_{<t}
\right)
=
\operatorname{softmax}
\left(
W_{\mathrm{LM}}
h_{q_t}^{(L)}
\right).
\end{gathered}
\label{eq:native_decoding}
\end{equation}

This design separates evidence compensation from lexical realization. The preceding routing module ensures that regional visual evidence re-enters the decision trajectory, whereas Native Semantic Recomposition converts this evidence into coherent token predictions through the model's pretrained generative capability.

\begin{table}[t]
\centering
\scriptsize
\setlength{\tabcolsep}{3pt}
\renewcommand{\arraystretch}{0.85}
\resizebox{0.48\textwidth}{!}{
\begin{tabular}{lcccccc}
\toprule
\multirow{2}{*}{\textbf{Method}}
& \multicolumn{2}{c}{\textbf{LLaVA-1.5-7B}}
& \multicolumn{2}{c}{\textbf{Qwen3-VL-4B}}
& \multicolumn{2}{c}{\textbf{InstructBLIP-7B}} \\
\cmidrule(lr){2-3}
\cmidrule(lr){4-5}
\cmidrule(lr){6-7}
& Acc.$\uparrow$ & F1$\uparrow$
& Acc.$\uparrow$ & F1$\uparrow$
& Acc.$\uparrow$ & F1$\uparrow$ \\
\midrule

\rowcolor{gray!15}
Origin
& 76.46 & 80.41
& 85.37 & 85.63
& 77.93 & 80.11 \\

VCD
& 77.79 & 80.70
& 89.11 & 89.23
& 80.93 & 81.47 \\

ICD
& 77.62 & 80.63
& 88.74 & 88.84
& 80.41 & 81.39 \\

CICD
& 78.69 & 81.51
& 88.96 & 88.97
& 81.59 & 83.09 \\

ClearSight
& 77.23 & 80.51
& 88.67 & 88.56
& 79.37 & 81.57 \\

VTI
& 81.21 & 82.87
& 85.33 & 85.40
& 79.52 & 81.20 \\

Nullu
& 80.63 & 82.62
& 87.74 & 87.09
& 76.78 & 80.34 \\

MemVR
& 81.24 & 82.14
& 89.32 & 88.95
& 81.85 & 81.63 \\

\rowcolor{gray!15}
EviAnchor
& \textbf{82.16} & \textbf{84.06}
& \textbf{90.93} & \textbf{91.41}
& \textbf{82.33} & \textbf{83.49} \\

\bottomrule
\end{tabular}
}
\caption{Performance evaluation on POPE (\%). Best results are highlighted in \textbf{bold}.}
\label{tab:pope_main}
\end{table}

\begin{table*}[t]
\centering
\scriptsize
\setlength{\tabcolsep}{2.5pt}
\renewcommand{\arraystretch}{0.85}
\resizebox{\textwidth}{!}{
\begin{tabular}{lcccccccccccc}
\toprule
\multirow{2}{*}{\textbf{Method}}
& \multicolumn{4}{c}{\textbf{LLaVA-1.5-7B}}
& \multicolumn{4}{c}{\textbf{Qwen3-VL-4B}}
& \multicolumn{4}{c}{\textbf{InstructBLIP-7B}} \\
\cmidrule(lr){2-5}
\cmidrule(lr){6-9}
\cmidrule(lr){10-13}
& CHAIRs$\downarrow$ & CHAIRi$\downarrow$ & Rec.$\uparrow$ & Len.
& CHAIRs$\downarrow$ & CHAIRi$\downarrow$ & Rec.$\uparrow$ & Len.
& CHAIRs$\downarrow$ & CHAIRi$\downarrow$ & Rec.$\uparrow$ & Len. \\
\midrule

\rowcolor{gray!15}
Origin
& 58.20 & 17.06 & 76.88 & 101.866
& 47.00 & 12.73 & 67.06 & 242.152
& 48.60 & 14.69 & 73.15 & 101.550 \\

VCD
& 57.00 & 17.70 & 76.88 & 102.116
& 48.40 & 10.93 & 67.68 & 241.214
& 57.20 & 16.00 & 72.53 & 96.824 \\

ICD
& 56.40 & 17.21 & 76.41 & 101.742
& 47.60 & 11.18 & 67.31 & 240.886
& 55.80 & 15.62 & 72.87 & 98.316 \\

CICD
& 48.80 & 14.84 & 75.57 & 102.406
& 48.40 & 10.89 & 67.06 & 240.054
& 48.00 & 13.06 & 71.35 & 99.642 \\

ClearSight
& 52.40 & 15.50 & 77.04 & 98.526
& 51.20 & 10.91 & 67.25 & 241.528
& 51.20 & 14.10 & \textbf{73.83} & 100.830 \\

VTI
& 50.40 & 15.06 & 57.18 & 97.854
& 43.00 & 10.15 & 62.46 & 242.888
& 40.20 & 12.06 & 62.71 & 127.654 \\

Nullu
& 49.40 & 15.40 & 74.46 & 95.928
& 43.20 & 13.71 & 54.69 & 218.512
& 40.60 & 12.36 & 67.99 & 76.738 \\

MemVR
& 49.20 & 14.65 & 77.20 & 100.842
& 43.60 & 10.48 & 67.47 & 241.026
& 40.60 & 12.38 & 71.94 & 99.284 \\

\rowcolor{gray!15}
EviAnchor
& \textbf{48.80} & \textbf{14.10} & \textbf{77.38} & 101.376
& \textbf{40.80} & \textbf{9.76} & \textbf{67.74} & 240.572
& \textbf{39.80} & \textbf{11.43} & 72.25 & 97.734 \\

\bottomrule
\end{tabular}
}
\caption{Performance evaluation on CHAIR (\%). CHAIRs and CHAIRi measure sentence-level and instance-level object hallucinations, respectively. Best results are highlighted in \textbf{bold}.}
\label{tab:chair_main}
\end{table*}

\section{Experiments}
\label{sec:experiments}

\subsection{Experimental Setup}
\label{sec:experimental_setup}

\subsubsection{Benchmarks and Metrics}
\label{sec:benchmarks}

To evaluate hallucinations at different levels of cognitive granularities, we adopt three representative benchmarks covering discriminative object-existence verification and open-ended generation.

\textbf{POPE}~\cite{li2023evaluating} evaluates object presence through binary yes-or-no questions under the \emph{Random}, \emph{Popular}, and \emph{Adversarial} settings, comprising 27,000 image--question pairs. We report accuracy, precision, recall, and F1 score.

\textbf{MMHal-Bench}~\cite{liu2024mitigating} contains 96 image--question pairs across eight categories and evaluates detailed free-form responses. Following the official protocol, GPT-4 assigns scores from 0 to 6 based on factual correctness and informativeness. We report the average score and hallucination rate, treating scores below 3 as hallucinated.

\textbf{CHAIR}~\cite{rohrbach2018object} measures unsupported object mentions in generated descriptions. $\mathrm{CHAIR}_{s}$ is the proportion of descriptions containing hallucinated objects, while $\mathrm{CHAIR}_{i}$ is the proportion of hallucinated object mentions. We evaluate on a set of 500 COCO val2014 images and additionally report object recall and average response length to account for overly short outputs.

\subsubsection{Implementation Details}
\label{sec:implementation_details}

We apply EviAnchor to three LVLMs with different architectures and model scales: LLaVA-1.5-7B~\cite{liu2023llava}, InstructBLIP-Vicuna-7B~\cite{dai2023instructblip}, and Qwen3-VL-4B-Instruct~\cite{Qwen3-VL}. All experiments use greedy decoding.

For the 32-layer LLaVA-1.5-7B and InstructBLIP-Vicuna-7B, we use layers 0--15, 16--23, and 24--31 for $\mathcal{L}{\mathrm{pool}}$, $\mathcal{L}{\mathrm{comp}}$, and $\mathcal{L}_{\mathrm{native}}$, respectively. For the 36-layer Qwen3-VL-4B-Instruct, the corresponding ranges are 0--17, 18--26, and 27--35.

We partition each image into a $3\times3$ spatial grid and insert $K=9$ REA slots, with each slot corresponding to one spatial region. The text-perturbation strength is set to $\lambda=0.3$. All experiments are conducted on NVIDIA A40 GPUs.

\subsubsection{Baselines}
\label{sec:baselines}

To comprehensively evaluate the effectiveness of EviAnchor, we compare it with five representative training-free hallucination mitigation methods. We divide the baselines into two categories according to where the intervention is applied: (i) Output-space methods, including VCD~\cite{leng2024mitigating}, ICD~\cite{wang2024mitigating}, and CICD~\cite{zhao2025crossimagecontrastivedecodingprecise}. (ii) Representation-space methods, including VTI~\cite{liu2025reducing}, ClearSight~\cite{yin2025clearsight}, Nullu~\cite{yang2025nullu}, and MemVR~\cite{zou2024look}.
All baselines are implemented using their officially recommended hyperparameters. We use identical input samples, decoding configurations, and evaluation protocols for all methods to ensure a fair comparison.

\begin{table}[t]
\centering
\footnotesize
\setlength{\tabcolsep}{4pt}
\renewcommand{\arraystretch}{0.95}
\begin{tabular}{lcccccc}
\toprule
\multirow{2}{*}{\textbf{Method}}
& \multicolumn{2}{c}{\textbf{LLaVA-1.5}}
& \multicolumn{2}{c}{\textbf{Qwen3-VL}}
& \multicolumn{2}{c}{\textbf{InstructBLIP}} \\
\cmidrule(lr){2-3}
\cmidrule(lr){4-5}
\cmidrule(lr){6-7}
& Score$\uparrow$ & Rate$\downarrow$
& Score$\uparrow$ & Rate$\downarrow$
& Score$\uparrow$ & Rate$\downarrow$ \\
\midrule

\rowcolor{gray!15}
Origin
& 2.64 & 58.3
& 4.04 & 40.6
& 2.06 & 64.4 \\

VCD
& 2.25 & 63.5
& 4.17 & 37.5
& 2.08 & 65.6 \\

ICD
& 2.31 & 57.3
& 4.26 & 36.5
& 2.18 & 61.5 \\

CICD
& 2.66 & 58.3
& 4.17 & 36.5
& 2.26 & 60.4 \\

ClearSight
& 2.52 & 57.3
& 4.00 & 39.6
& 2.12 & 63.5 \\

VTI
& 2.72 & 51.0
& 4.25 & 36.5
& 2.26 & 59.4 \\

Nullu
& 2.70 & 54.2
& 4.22 & 39.6
& 2.30 & \textbf{56.2} \\

MemVR
& 2.71 & 51.0
& 4.23 & 35.4
& 2.24 & 57.3 \\

\rowcolor{gray!15}
EviAnchor
& \textbf{2.78} & \textbf{49.0}
& \textbf{4.30} & \textbf{29.2}
& \textbf{2.34} & \textbf{56.2} \\

\bottomrule
\end{tabular}
\caption{Performance evaluation on MMHal-Bench. Best results for each backbone are highlighted in \textbf{bold}.}
\label{tab:mmhal_main}
\end{table}

\subsection{Main Results}

\paragraph{POPE Results}
As shown in Table~\ref{tab:pope_main}, EviAnchor achieves the highest accuracy and F1 score on all three backbones, demonstrating consistent improvements in object-existence verification. Compared with Origin, EviAnchor improves accuracy/F1 by 5.70/3.65 percentage points on LLaVA-1.5, 5.56/5.78 points on Qwen3-VL, and 4.40/3.38 points on InstructBLIP. Although ClearSight and Nullu achieve high recall on some backbones, their substantially lower precision indicates a tendency to over-predict the answer \texttt{Yes} excessively. In contrast, EviAnchor maintains a better balance between precision and recall. The improvement is particularly pronounced on Qwen3-VL, where EviAnchor achieves 90.93 accuracy and 91.41 F1. These results demonstrate that regional evidence compensation improves visual discrimination without relying on a positive-response bias.

\paragraph{CHAIR Results}
Table~\ref{tab:chair_main} shows that EviAnchor consistently reduces object hallucinations relative to Origin. On LLaVA-1.5, it reduces $\mathrm{CHAIR}_{s}$ from 58.20 to 48.80 and $\mathrm{CHAIR}_{i}$ from 17.06 to 14.10, while preserving the highest recall of 77.38. On Qwen3-VL, EviAnchor achieves the best $\mathrm{CHAIR}_{s}$, $\mathrm{CHAIR}_{i}$, and recall scores of 40.80, 9.76, and 67.74, respectively. On InstructBLIP, it reduces the two hallucination metrics from 48.60/14.69 to 39.80/11.43, while retaining a recall of 72.25. Moreover, its average response lengths remain close to those of Origin across all backbones, indicating that the improvements do not result from trivially generating shorter descriptions.

\paragraph{MMHal-Bench Results}
As reported in Table~\ref{tab:mmhal_main}, EviAnchor achieves the best performance across all three backbones. On LLaVA-1.5, it increases the average score from 2.64 to 2.78 and decreases the hallucination rate from 58.3 to 49.0. On Qwen3-VL, it improves the score from 4.04 to 4.30 while substantially reducing the hallucination rate from 40.6 to 29.2. On InstructBLIP, EviAnchor increases the score from 2.06 to 2.34 and reduces the hallucination rate from 64.4 to 56.2. Compared with the strongest baseline on LLaVA-1.5 and Qwen3-VL, EviAnchor further reduces the hallucination rate by 2.0 and 7.3 percentage points, respectively.

Overall, the results demonstrate that EviAnchor generalizes across model architectures and improves visual grounding in both discriminative and open-ended generation tasks.

\subsection{Ablation Studies}

We conduct module ablations on LLaVA-1.5-7B using fixed subsets of POPE and CHAIR. 

\subsubsection{Module Ablation}
\label{sec:module_ablation}

\begin{table}[t]
\centering
\scriptsize
\setlength{\tabcolsep}{2pt}
\renewcommand{\arraystretch}{0.85}
\resizebox{\columnwidth}{!}{
\begin{tabular}{lccccc}
\toprule
\multirow{2}{*}{\textbf{Variant}}
& \multicolumn{2}{c}{\textbf{POPE}}
& \multicolumn{3}{c}{\textbf{CHAIR}} \\
\cmidrule(lr){2-3}
\cmidrule(lr){4-6}
& Acc.$\uparrow$
& F1$\uparrow$
& CHAIR$_s\downarrow$
& CHAIR$_i\downarrow$
& Rec.$\uparrow$ \\
\midrule

\rowcolor{gray!15}
Origin
& 84.73
& 84.21
& 58.20
& 17.06
& 76.88 \\

w/o Regional Anchoring
& 85.27
& 85.43
& 53.60
& 15.14
& 76.95 \\

w/o Evidence Routing
& 85.57
& 85.83
& 52.20
& 14.78
& 76.90 \\

w/o Native Recomposition
& 85.60
& 85.34
& 51.40
& 14.52
& 71.80 \\

\rowcolor{gray!15}
EviAnchor
& \textbf{86.37}
& \textbf{86.76}
& \textbf{48.80}
& \textbf{14.10}
& \textbf{77.38} \\

\bottomrule
\end{tabular}
}
\caption{Component ablation results on POPE and CHAIR. 
\emph{w/o Regional Anchoring} removes the REA slots and redistributes the transferred attention mass to the original visual tokens. 
\emph{w/o Evidence Routing} disables textual perturbation and attention transfer. 
\emph{w/o Native Recomposition} extends the routing intervention to the final decoder layer.}
\label{tab:component_ablation}
\end{table}

As shown in Table~\ref{tab:component_ablation}, EviAnchor achieves the best performance across all metrics. 
Removing regional anchoring decreases POPE F1 by 1.33 points and increases CHAIR$_s$ from 48.80 to 53.60, confirming the benefit of compact regional evidence representations. Removing evidence routing further increases CHAIR$_s$ to 52.20, indicating that preserved visual anchors alone do not guarantee their effective utilization by later decision states. Extending the routing intervention to the final layer reduces recall from 77.38 to 71.80, highlighting the importance of native upper-layer recomposition for maintaining response completeness.

\begin{table}[t]
\centering
\scriptsize
\setlength{\tabcolsep}{1pt}
\renewcommand{\arraystretch}{0.85}
\resizebox{\columnwidth}{!}{
\begin{tabular}{lcccccc}
\toprule
\multirow{2}{*}{\textbf{REA Design}}
& \multicolumn{2}{c}{\textbf{POPE}}
& \multicolumn{4}{c}{\textbf{CHAIR}} \\
\cmidrule(lr){2-3}
\cmidrule(lr){4-7}
& Acc.$\uparrow$
& F1$\uparrow$
& CHAIR$_s\downarrow$
& CHAIR$_i\downarrow$
& Rec.$\uparrow$
& Empty$\downarrow$ \\
\midrule

Global $1\times1$ (1)
& 85.10
& 85.71
& 54.80
& 15.64
& 76.70
& 0 \\

Spatial $2\times2$ (4)
& 85.10
& 85.80
& 52.60
& 15.06
& 76.90
& 0 \\

\rowcolor{gray!15}
Spatial $3\times3$ (9)
& 86.37
& 86.76
& 48.80
& 14.10
& \textbf{77.38}
& 0 \\

Spatial $4\times4$ (16)
& 87.43
& 87.28
& 49.60
& 10.90
& 71.48
& \textbf{8} \\

Spatial $4\times4$ (16)$^{\dagger}$
& 86.04
& 86.37
& 45.20
& 13.28
& 77.26
& 0 \\

Spatial $5\times5$ (25)
& \textbf{87.77}
& \textbf{87.35}
& 53.80
& \textbf{9.27}
& 54.98
& \textbf{25} \\

Spatial $5\times5$ (25)$^{\ddagger}$
& 87.21
& 86.88
& \textbf{44.00}
& 10.80
& 64.26
& \textbf{2} \\

\bottomrule
\end{tabular}
}
\caption{Ablation on the number of REA slots. Variants marked with
$\dagger$ and $\ddagger$ use relaxation strengths $\lambda=0.1$ and
$\lambda=0$, respectively.}
\label{tab:rea_granularity}
\end{table}

\subsubsection{REA Slot Granularity}
\label{sec:rea_granularity}

We further investigate the influence of the number of REA slots using evidence banks containing 1, 4, 9, 16, and 25 anchors, while keeping the remaining inference settings unchanged. As shown in Table~\ref{tab:rea_granularity}, increasing the number of anchors generally improves POPE performance, suggesting that finer evidence grouping benefits visual discrimination. However, excessively fine-grained evidence banks introduce generation instability. Under the default relaxation strength, the 16- and 25-slot configurations produce 8 and 25 empty responses, respectively, leading to substantial recall degradation. Their lower CHAIR$_i$ values are therefore partly influenced by reduced output coverage rather than purely improved hallucination suppression.

We further reduce the relaxation strength for larger evidence banks. For the 16-slot configuration, setting $\lambda=0.1$ eliminates empty responses and recovers recall from 71.48 to 77.26, while maintaining improved hallucination reduction. For the 25-slot configuration, removing relaxation decreases empty responses from 25 to 2 and improves recall from 54.98 to 64.26. However, its recall remains substantially lower than that of the $3\times3$ configuration, suggesting that an excessively large REA bank can interfere with normal generation even without textual relaxation. These results indicate that larger evidence banks provide stronger visual constraints but require careful control of intervention intensity. Accordingly, we adopt the $3\times3$ configuration as the default setting, which provides the best balance between visual grounding and response completeness.

\begin{figure}[t]
    \centering
    \scalebox{1}[0.95]{%
        \includegraphics[width=\linewidth]{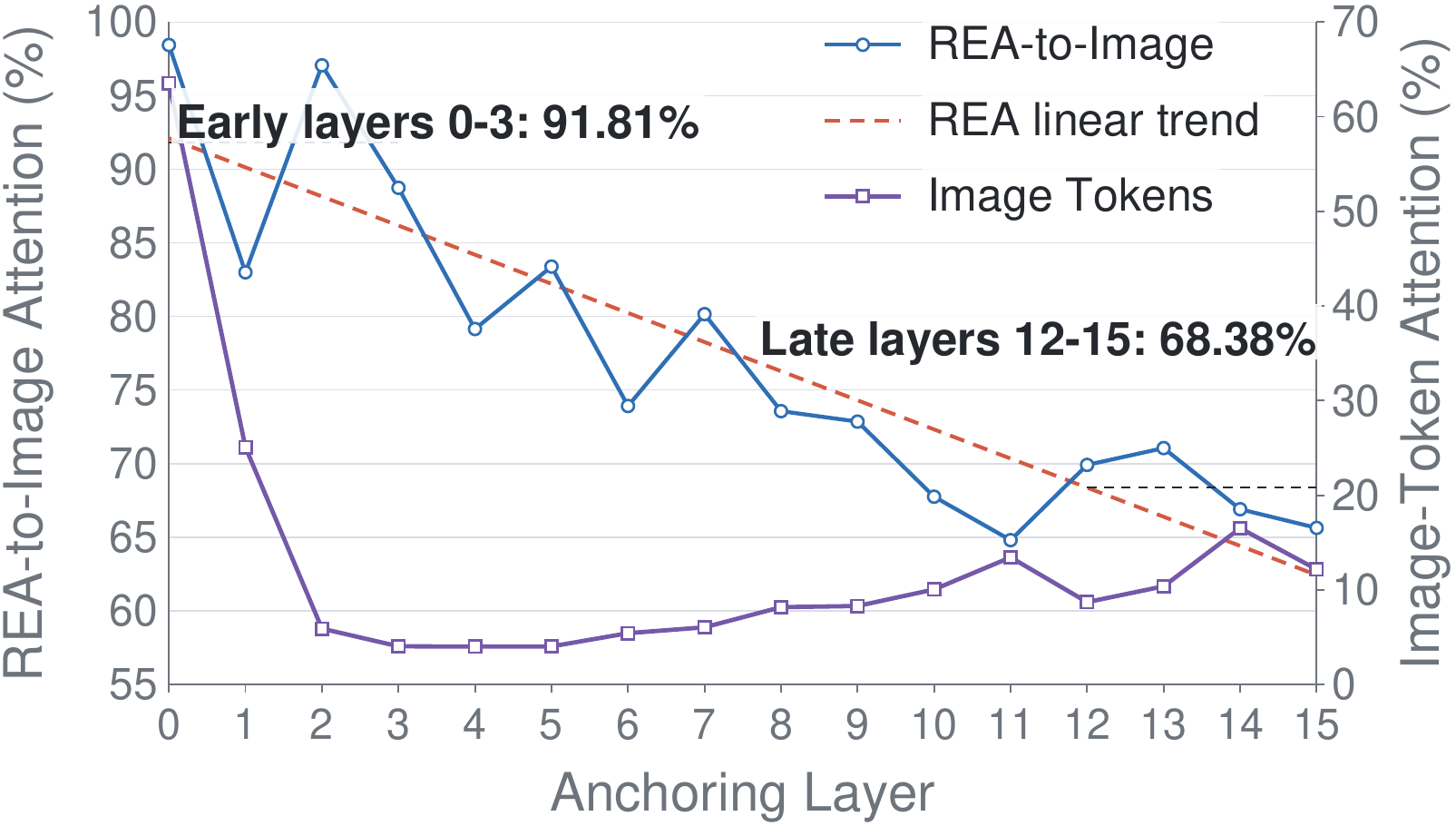}
    }
    \caption{
    Progressive visual evidence aggregation during regional anchoring. The REA-to-Image curve shows the attention from REA slots to their corresponding image tokens, while the Image Tokens curve shows the current decision state's attention to the original visual tokens.
    }
    \label{fig:progressive_evidence_aggregation}
\end{figure}

\begin{figure}[t]
    \centering
    \scalebox{1}[0.88]{%
        \includegraphics[width=\linewidth]{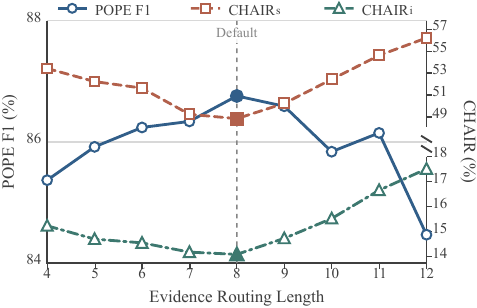}
    }
    \caption{Effect of evidence-routing length.}
    \label{fig:routing_length}
\end{figure}

\subsubsection{Module Transition Boundaries}

\label{sec:routing_length}

To motivate the boundary of Regional Evidence Anchoring, we examine how strongly the REA slots attend to their corresponding image tokens across the lower decoder layers. As shown in Figure~\ref{fig:progressive_evidence_aggregation}, REA-to-image attention exhibits an overall decreasing trend. The average attention is 91.81\% in the early layers (0--3), compared with 68.38\% in the late anchoring layers (12--15). This pattern is consistent with progressive evidence aggregation: the earliest REA states rely most strongly on direct access to image tokens, whereas later states increasingly operate on evidence that has already been accumulated.
Together with Figure~\ref{fig:layerwise_visual_analysis}, which shows that visual influence is concentrated in the first half of the decoder and weakens thereafter, this result provides mechanistic support for assigning layers 0--15 to Regional Evidence Anchoring.

We then vary the duration of the Decision-Conditioned Evidence Routing from 4 to 12 layers, with the remaining upper layers assigned to Native Semantic Recomposition.
As shown in Figure~\ref{fig:routing_length}, extending evidence routing from 4 to 8 layers consistently improves visual grounding: POPE F1 increases from 85.37 to 86.76, while CHAIR$_s$ and CHAIR$_i$ decrease from 53.20/15.35 to 48.80/14.11. Further extending the routing stage generally degrades all three metrics at 12 layers. These results indicate that a short routing stage is insufficient for the preserved evidence to influence subsequent decisions, whereas an overly long routing stage excessively suppresses attention to the target text and leaves inadequate upper-layer computation for native semantic recomposition, thereby disrupting reliable generation. The eight-layer configuration therefore provides the best balance between visual evidence compensation and textual-semantic preservation.

\section{Conclusion}
\label{sec:conclusion}

In this work, we investigated multimodal hallucination from the perspective of cross-layer visual evidence utilization. Our layer-wise analyses show that visual evidence is primarily incorporated into answer-side representations in early-to-middle decoder layers, while its direct influence weakens in later layers. Based on this observation, we proposed EviAnchor, a training-free and single-branch inference framework that preserves regional visual evidence in REA slots, reconnects it with the current generation state through decision-conditioned routing, and restores native upper-layer computation for coherent response generation. Experiments across multiple benchmarks validate the effectiveness of cross-layer visual evidence preservation and reuse for improving the reliability of LVLM generation.

\bibliography{aaai2027}


\end{document}